\documentclass{article}

\usepackage{arxiv}

\usepackage[utf8]{inputenc} 
\usepackage[T1]{fontenc}    
\usepackage{hyperref}       
\usepackage{url}            
\usepackage{booktabs}       
\usepackage{amsfonts}       
\usepackage{nicefrac}       
\usepackage{microtype}      
\usepackage{lipsum}
\usepackage{graphicx}
\usepackage[inline]{enumitem}
\usepackage{multirow}

\newcommand{\emphb}[1]{\emph{\textbf{#1}}}
\newcommand\Tstrut{\rule{0pt}{2.6ex}}         
\newcommand\Bstrut{\rule[-0.9ex]{0pt}{0pt}}   %

\title{Learning to aggregate feature representations}

\author{
  Guy Gaziv \\
  Dept. of Computer Science and Applied Math \\
  The Weizmann Institute of Science \\
  76100 Rehovot, Israel \\
  \texttt{guy.gaziv@weizmann.ac.il} \\
}

\begin{document}
\maketitle

\begin{abstract}
The Algonauts challenge requires to construct a multi-subject encoder of images to brain activity. Deep networks such as ResNet-50 and AlexNet trained for image classification are known to produce feature representations along their intermediate stages which closely mimic the visual hierarchy. However the challenges introduced in the Algonauts project, including combining data from multiple subjects, relying on very few similarity data points, solving for various ROIs, and multi-modality, require devising a flexible framework which can efficiently accommodate them. Here we build upon a recent state-of-the-art classification network (SE-ResNeXt-50) and construct an adaptive combination of its intermediate representations. While the pretrained network serves as a backbone of our model, we \emph{learn} how to aggregate feature representations along five stages of the network. During learning, our method enables to modulate and screen outputs from each stage along the network as governed by the optimized objective. We applied our method to the Algonauts2019 fMRI and MEG challenges. Using the combined fMRI and MEG data, our approach was rated among the leading five for both challenges. Surprisingly we find that for both the lower and higher order areas (EVC and IT) the adaptive aggregation favors features stemming at later stages of the network.
\end{abstract}

\section{Introduction}
Encoding seen images into corresponding elicited brain activity of human subjects is an active field of research.

Deep networks trained for image classification give rise to a valuable feature representation. This representation have been shown to closely mimic the hierarchy of the human visual system~\cite{Guclu2015DeepStream, Horikawa2017HierarchicalFeatures, Wen2018DeepCategorization, Grossman2018DeepRepresentation, Shen2018End-to-endActivity, Eickenberg2017SeeingSystem, Richard2018OptimizingActivity}.

The Algonauts fMRI (or MEG) challenge tasks to design an image encoder that transforms pixel values to an \emph{`fMRI-like' (or `MEG-like')} embedding~\cite{CichyTheIntelligence}. Here, `fMRI-like' refers to the sense of Representational Similarity Analysis (RSA): A good encoder is one for which pairwise similarity of images computed in their predicted embedding space is close to that computed in the fMRI space using their underlying fMRI recordings~\cite{Kriegeskorte2008RepresentationalNeuroscience}.

We categorize the posed challenges as follows and detail on each thereafter:
\begin{itemize}
    \item Cross subject generalization.
    \item Region Of Interest (ROI) specificity (for fMRI, or interval-time specificity for MEG).
    \item Poor instructive signal.
    \item Limited data.
    \item Multi-modality (fMRI and MEG).
\end{itemize}
 

Three sets of images are considered (92, 118, and 78) for 15 subjects. The first two sets are designated for training and are provided with their corresponding Representational Dissimilarity Matrix (RDM) for each subject. The last set of images is the held-out (test) data for which a predicted RDM is desired. Importantly, \textbf{the predicted RDM is not subject-specific} albeit is required to correlate well against ground truth RDMs of \emph{all} of the 15 subjects. 

The entire fMRI challenge is instantiated for \textbf{two distinct brain regions}, hallmarks of the two extremes of the visual processing: early visual and inferior temporal cortices (EVC and IT). Analogously for MEG, \textbf{two time-intervals}, early and late, are considered as two distinct challenges. 

The training data provides only a \textbf{weakly informative instructive signal} for supervised learning. Every pair of images is associated with as little as a single number per subject, which represents their similarity level. On the other hand upgrading to voxel-level data from \cite{Cichy2016Similarity-BasedRecognition} (or using other external datasets), is non-trivial due to lack of common ground across subjects and acquisition parameters of different datasets. 

Moreover the number of images or the number of pairwise combinations which constitute the \textbf{training data are very small} compared to commonly used training datasets in Deep Learning - Only about 200 distinct images are considered, labeled with approximately 10K distinct pairwise similarity values per subject for the entire training data. These numbers cannot span the space of natural images, or the space of all their possible pairwise similarities.  

Attempting to increase dataset size with MEG-based RDMs introduces the \textbf{multi-modality challenge}: For example, determining correspondences of EVC/IT to MEG phases.

\begin{figure}[t] 
\centering
\includegraphics{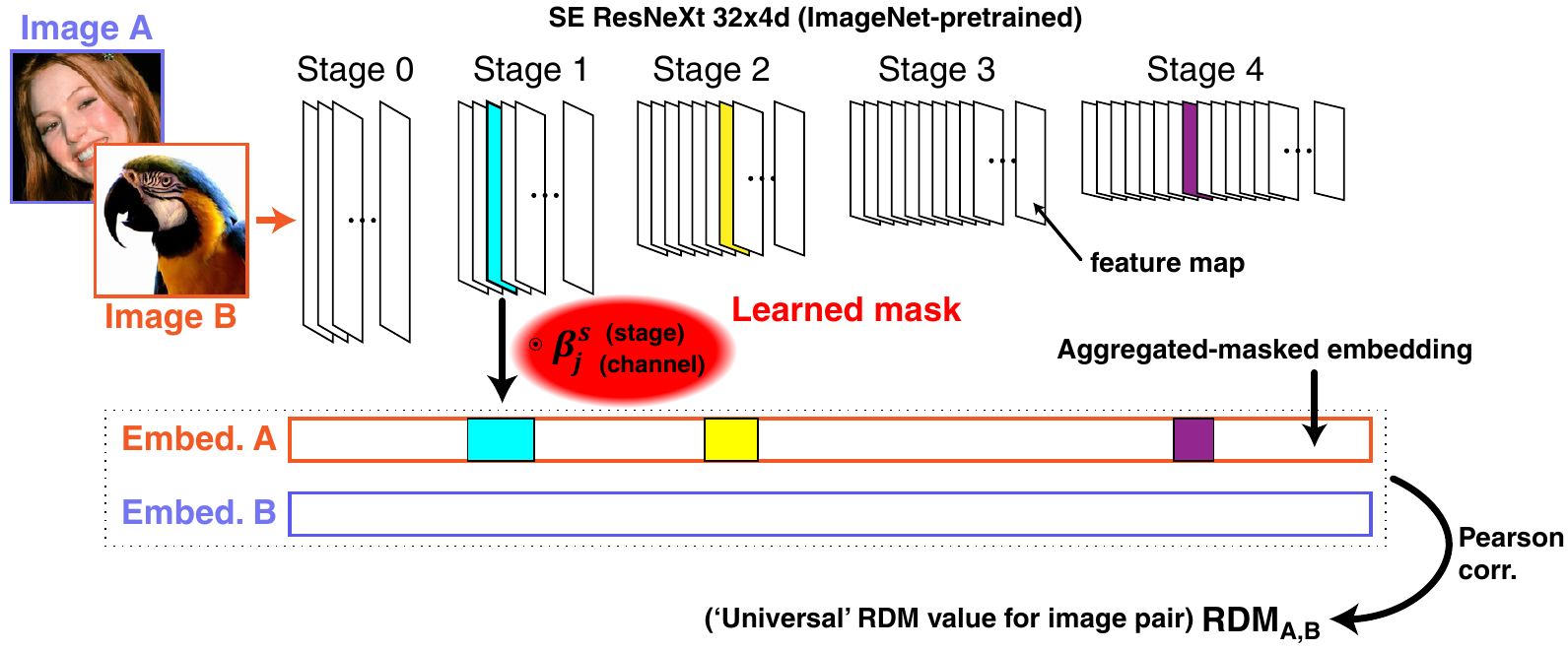}
\caption{ \textbf{Illustration of our `Learning to aggregate' proposed method.} Features arising from stage $s$ and feature map (channel) $j$ are modulated (scaled) by their corresponding learned coefficient $\beta_{j}^{s}$. All modulated features are concatenated to form the image embedding. The embeddings of every pair of images are Pearson-correlated to yield the corresponding `universal' (i.e., good for all 15 subjects) RDM value. 
}
\label{fig:Method}
\end{figure}

Here we show customization of a state-of-the-art deep network trained for image classification to meet the challenge's demands. Fig~\ref{fig:Method} shows the proposed approach. Our method relies on a pretrained deep network for image classification. At the core of our method we aggregate over the features produced at various stages along the network as our multi-level embedding (see `Hypercolumn'~\cite{Hariharan2015HypercolumnsLocalization}). Then, we \emphb{parameterize the outputs of the various stages by a continuous mask}. This mask, initialized to identity at every stage, is \emphb{learned} in a supervised way. It enables to modulate and screen outputs from each stage along the network according to its importance in reconstructing the training data RDMs. Importantly, the activity at various ROIs along the visual hierarchy have been shown to match the hierarchy of deep networks~\cite{Wen2018DeepCategorization, Guclu2015DeepStream}, however the exact hierarchical correspondence remains vague. Thus our `Learning to aggregate' approach naturally accommodates application to arbitrary ROI, including the targeted EVC and IT: It flexibly learns to harness the representations at the various stages according to their utility for the task. 

Here we focused on the Squeeze-and-Excitation variant of ResNet-50, specifically SE-ResNeXt-50~(32x4d)~\cite{HuSqueeze-and-ExcitationNetworks}, as the pretrained network of choice. While ResNet50 is to-date considered to yield the most matching feature representation to brain activity~\cite{Wen2018DeepCategorization}, this architecture has shown improved image-classification performance over ResNet-50 (and a few other variants) at comparable parameter count and computational cost~\cite{Bianco2018BenchmarkArchitectures}. 

We found that training on RDMs from both fMRI and MEG training data improves the results in some cases.

\section{Method}

To formulate the supervised learning problem given images and RDM data, we considered image pairwise combinations and their corresponding dissimilarity entry. For each image-pair the two embedding vectors were Pearson-correlated to produce the RDM value (Fig \ref{fig:Method}).

We used the SE-ResNeXt-50~(32x4d) pytorch implementation initialized to its ImageNet pretrained weights~\cite{HuSqueeze-and-ExcitationNetworks}. This served as the major backbone of our method while the poor and sparse RDM data was used as a moderate fine-tune atop. This finetune either targeted all model parameters (backbone + mask) or merely the mask (see Results).  

\textbf{The training data} which we used considered the provided 92- and 118-image datasets and their corresponding RDMs for all subjects, and from both fMRI and MEG training data. The total of four datasets were concatenated to form a single dataset. The resulting dataset was randomly split to training and validation at a ratio of 90:10. We associated the early and late time-intervals of the MEG with EVC and IT ROIs of the fMRI RDMs respectively.

\subsection{Mask resolution}
We experimented with 5 or 17 intermediate tracked stages, and with various degrees of mask resolution: single multiplicative scalar per stage, one per channel within a stage, and down to one per single feature (stage, channel and spatial location). We focused on our simplest and well performing configuration which was due to 5 stages, `layer0'-`layer4' where in each stage the channels are parameterized separately (but space remains uniformly scaled). All masks are initialized to identity at the training onset.

\subsection{Cross subject disparity}
Since we are aiming at a single encoder for all subjects, the disparity across them given the \emph{same} image-pair has a detrimental effect on the training. Hence we defined this variance as our noise term and used it to weigh entries' contribution accordingly during optimization. Let $N$ denote the noise, given by the RDM standard deviation across the subjects. Then the reliability-weight on the RDM reconstruction loss for a given image-pair entry $\{i,j\}$ reads,
\begin{equation}
    w_{i,j}=\left(\frac{1}{N+\alpha\bar{N}}\right)^{\beta},
\end{equation}
where $\alpha=0.25$, $\beta=1$, and $\bar{N}$ is the mean of the noise.

Importantly, while we predict a single RDM value given an image-pair, we simultaneously constrain it to reconstruct all the 15 values within the training data, corresponding to all subjects. This reflects the \emph{universality} property which is desired from the encoder (at least in the sense of the 15 subjects).
\subsection{Implementation details}
\textbf{Data augmentation} was practiced to enrich the training data. Specifically we used random crop over the input images of 87.5\% of input size. Additional variants of data augmentation included Gaussian random sampling of timestamp within the MEG RDM values\footnote{Drawing samples at random throughout the recorded time interval with increased probability toward the interval's middle point and diminishing toward the endings.}.

\textbf{Training} duration varied 10-40 epochs mostly depending on whether the entire network was finetuned (EVC) or just the mask module (IT). We used Adam optimizer with initial learning rate of 0.01, and batch size of 40. 

\textbf{The loss} applied to predicted RDMs was $L_{1}$ (MAE) reconstruction loss, however MSE appeared to yield comparable results.

\section{Results}
We achieved scores of 20.21\% and 42.26\% noise-normalized average $R^{2}$ for fMRI and MEG challenged respectively. This was achieved using our method with five stages of aggregation and a single multiplicative parameter for each feature map (channel). The specific best performing configurations of our method varied depending on the specific sub-challenge category. Table~\ref{table:Summary} shows a summary of our final performance and the corresponding configurations for fMRI and MEG.

\begin{table}[h!]
\centering
\begin{tabular}{ |c|c|c|c| } 
\hline 
\textbf{Category} & \textbf{Noise-Normalized} $R^{2}~(\%)$ & \textbf{Configuration} \Tstrut\Bstrut\\ 
\hline
fMRI-EVC & 24.93 & fMRI + MEG data, MEG Gaussian sampling, train all, using SNR weights, 15 epochs \Tstrut\Bstrut\\ 
fMRI-IT & 15.55 & fMRI data only, train mask only, no SNR weights, 15 epochs \Tstrut\Bstrut\\ 
\hline
MEG-early & 51.21 & \multirow{2}{*}{MEG data only, MEG Gaussian sampling, train all, using SNR weights, 40 epochs} \Tstrut\Bstrut\\ 
MEG-late & 35.10 &  \Tstrut\Bstrut\\ 
\hline
\end{tabular} \\ [0.5ex]
\caption{\textbf{Final performance and configuration summary.}}
\label{table:Summary}
\end{table}

\begin{figure}[ht] 
\centering
\includegraphics{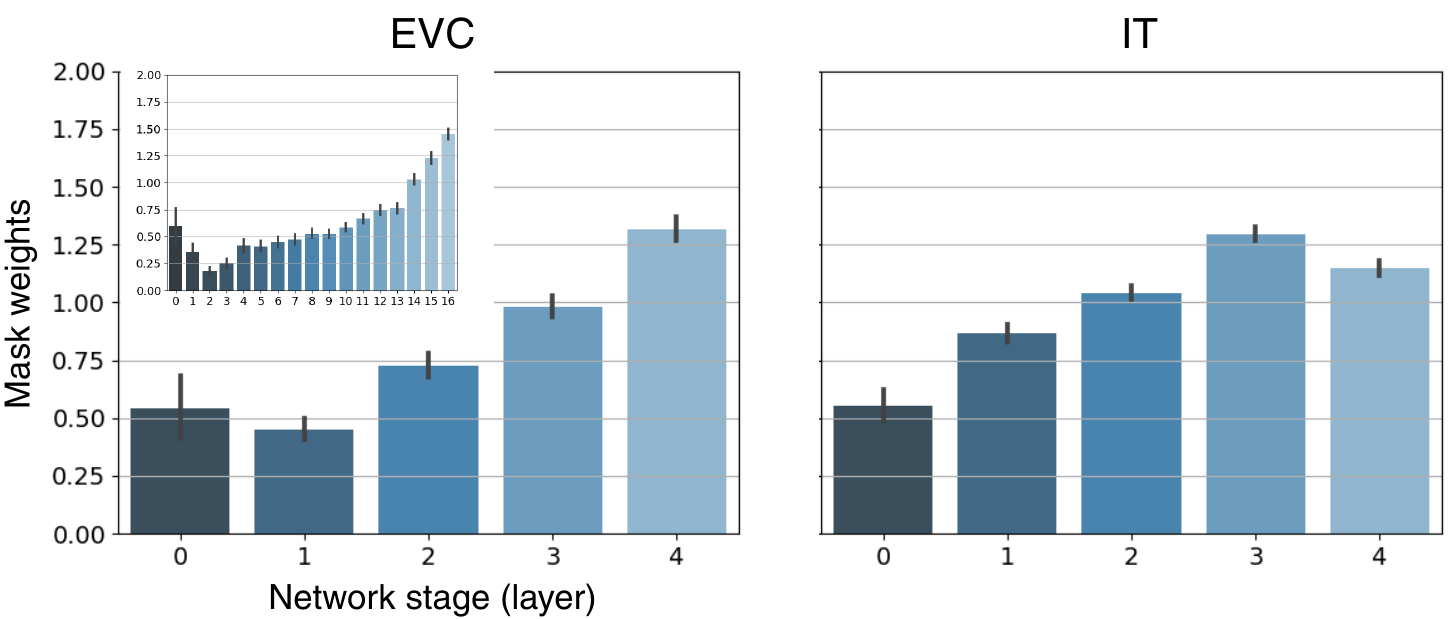}
\caption{ \textbf{Tuning of network stages at end of training for EVC and IT.} 
Both EVC and IT favor tuning to higher stages of the network. The inset shows corresponding scenario for the case where 17 network stages contribute to representation (with a mask) for the EVC case; Even under the finer stage-resolution EVC favors higher layers. Bars show the mean weight across the weights assigned to each channel; Error bars indicate 95\% CI. 
}
\label{fig:StageTuning}
\end{figure}

Fig~\ref{fig:StageTuning} shows the resulting tuning to network stages post training for EVC and IT. Surprisingly, no particular preference towards early stages of the network was recorded for the EVC-based model. We note that this result was based solely on fMRI data (either EVC or IT) and did not consider the MEG data.

We found that using a combination of RDMs from fMRI and MEG training data confers a leap improvement to our results for fMRI-EVC challenge. On the other hand, for any of the MEG challenges, best results were rather accomplished when fMRI data were discarded.

We found that the fMRI-EVC and both MEG challenges benefit from finetuning the entire network (i.e., the mask and the pretrained weights). For fMRI-IT, however, best results were achieved when only the mask was trained while the pretrained backbone has remained fixed. Optimizing the entire network for this challenge dramatically degraded the results.

The inclusion of reliability-weights had a positive impact on our results for the most part, albeit only to a moderate degree.

\section{Conclusion}
We present a method for learning to aggregate features along deep networks representations in a supervised setting.
We framed the Algonauts2019 fMRI and MEG challenges as a supervised learning problem given pairs of images and their pairwise similarity as reconstruction target.

we applied our method under this setting and optimized fMRI- and MEG-like encoders for two distinct ROIs from the visual cortex (fMRI) and for two time intervals (MEG). 

We report competitive performance using our method, which is rated among the five leading solutions for both fMRI and MEG challenges of the Algonauts project.

\section*{Acknowledgments}
This project has received funding from the European Research Council (\textbf{ERC}) under the European Union’s Horizon 2020 research and innovation programme (grant agreement No 788535).

\bibliographystyle{unsrt}

\end{document}